\documentclass[twoside]{article}

\usepackage[accepted]{aistats2021}
\usepackage[round]{natbib}

\usepackage{graphicx}
\usepackage{tabularx}
\usepackage{amssymb}
\usepackage{amsmath}
\newcommand{\vect}[1]{\boldsymbol{#1}}
\usepackage{enumitem}
\usepackage[ruled]{algorithm2e}

\usepackage[font=small,labelfont=bf]{caption}
\begin{document}

%
\runningtitle{Latent-Graph Convolutional Networks}

%
\runningauthor{Floris A.W. Hermsen, Peter Bloem, Fabian Jansen \& Wolf B.W. Vos }

\twocolumn[

\aistatstitle{End-to-End Learning from Complex Multigraphs \\ with Latent-Graph Convolutional Networks}

\aistatsauthor{ Floris A.W. Hermsen \And Peter Bloem}

\aistatsaddress{ University of Amsterdam \& Owlin \And VU Amsterdam}

\aistatsauthor{ Fabian Jansen \And Wolf B.W. Vos }

\aistatsaddress{ ING WBAA \And Heineken }
]

\begin{abstract}
We study the problem of end-to-end learning from complex multigraphs with potentially very large numbers of edges between two vertices, each edge labeled with rich information. Examples range from communication networks to flights between airports or financial transaction graphs. We propose Latent-Graph Convolutional Networks (L-GCNs), which propagate information from these complex edges to a latent adjacency tensor, after which further downstream tasks can be performed, such as node classification. We evaluate the performance of several variations of the model on two synthetic datasets simulating fraud in financial transaction networks, ensuring the model must make use of edge labels in order to achieve good classification performance. We find that allowing for nonlinear interactions on a per-neighbor basis boosts performance significantly, while showing promising results in an inductive setting. Finally, we demonstrate the use of L-GCNs on real-world data in the form of an urban transportation network.
\end{abstract}

\section{INTRODUCTION}
Much of human interaction can be expressed as large, complex network structures. Examples vary from communication and transportation networks to financial transactions and knowledge graphs. Valuable information resides in these networks, and detecting the right patterns may have benefits ranging from more efficient logistics all the way to crime prevention and fraud detection.

In recent years much progress has been made through the development of \emph{Graph Convolutional Networks} (GCNs) \citep{KipfW16}. These architectures and their spin-offs, such as \emph{Relational Graph Convolutional Networks} (R-GCNs) \citep{RGCN_paper}, have shown great performance in node classification and link prediction tasks, improving state-of-the-art results. Until now, these networks have been of a relatively straightforward structure with discrete relations, such as scientific citation networks or simple social networks.

In reality however, we often encounter graphs with many additional degrees of complexity. E.g. financial transaction networks, which are multigraphs with edges representing rich information (sequences of transactions). Labels are usually not available at the edge level and one often does not know which patterns one is looking for. As a solution, we look to deep models that can be trained end-to-end: first deriving representations of the relationships between nodes, and then using these in a node classification model. This has yet to be demonstrated on data sets of this complexity within the domain of GCNs.

In this study, we investigate if we can perform end-to-end learning on multigraphs in the form of two synthetic financial transaction networks as well as a real-world urban transportation network. We propose the concept of \emph{Latent-Graph Convolutional Networks} (L-GCNs), in which learning mechanisms transform multi-edges into latent representations in the form of an adjacency tensor, which then serves as input to a GCN-like architecture for further propagation. We hypothesize that L-GCNs can exploit information unavailable to either non-GCN architectures relying on simple local neighborhood aggregation or standard GCN architectures that do not have access to edge-level information.

We start with two synthetic datasets we generate ourselves in order to be certain that important information resides at the transaction level, and hence can only be extracted by means of successful end-to-end learning. This furthermore provides a controlled setting in which we can experiment with different data patterns and examine the performance of our models in both transductive and inductive settings. We conclude by demonstrating the use of L-GCN on real-world data in the form of an urban transportation network.

Even though these kinds of highly complex networks are omnipresent, we address a novel problem domain and publicly available datasets do not exist, also due to security and privacy concerns. We will therefore share the data sets we have created with the community so they may serve as benchmarks\footnote{Our data sets and implementations can be found on GitHub: github.com/florishermsen/L-GCN}.

\section{RELATED WORK}

\subsection{Graph Networks \& Convolutions}
\label{subsection:GCN}

Graph neural networks \citep{zhou2018graph,zhang2018deep} are a recent innovation in machine learning, where neural networks take their topology from the graph structure present in the data (either explicitly or latently). The most common graph neural networks are probably \emph{Graph Convolutional Networks} (GCNs) \citep{BrunaZSL13,DuvenaudMAGHAA15,DefferrardBV16,KipfW16}. \citeauthor{KipfW16} use spectral graph theory to motivate a neural network architecture with the following propagation rule:
\begin{equation}
    \vect{H}^{(l+1)} = \sigma \big(\tilde{\vect{D}}^{-\frac{1}{2}}\tilde{\vect{A}}\tilde{\vect{D}}^{-\frac{1}{2}}\vect{H}^{(l)}\vect{W}^{(l)}\big),
    \label{eq:GCN_architecture}
\end{equation}
computing a next layer of node embeddings $\vect{H}^{(l+1)}$ from the previous layer $\vect{H}^{(l)}$. Given a graph $\mathcal{G}$ with vertices $v_i \in \mathcal{V}$, $\tilde{\vect{A}}=\vect{A}+\vect{I}_\mathcal{V}$ is the adjacency matrix of shape $|\mathcal{V}|\times|\mathcal{V}|$ with added self-connections\footnote{Self-connections are added in order to allow vertices to retain their original properties to an extent. $\vect{I}_\mathcal{V}$ is the identity matrix of size $|\mathcal{V}|$.} and $\tilde{\vect{D}}_{i}=\sum_{j}\tilde{\vect{A}}_{ij}$ represents a normalization matrix obtained through row and column-wise summation of $\tilde{\vect{A}}$\footnote{For directed graphs, only row-wise normalization applies.}. By convention, $\vect{H}^{(0)} = \vect{X}$, which is the original $|\mathcal{V}|\times|\mathcal{F}|$ matrix representing the graph vertices $\mathcal{V}$ and their features $\mathcal{F}$. Finally, $\sigma$ is a nonlinear activation and $\vect{W}^{(l)} \in \mathbb{R}^{|\mathcal{F}^{(l)}|\times|\mathcal{F}^{(l+1)}|}$ is the layer-specific weight matrix, which determines the size of the new embedding. The final embedding layer $\vect{H}^{(M)}$ can serve as an output layer of appropriate size, depending on the downstream task at hand, e.g. node classification.

\subsubsection{Message Passing Frameworks}
\label{subsubsection:MPF}

Equation \ref{eq:GCN_architecture} can be reformulated as a special case of a message-passing framework (MPF), which defines propagation rules at the node level \citep{GilmerSRVD17}. 
In case of a binary adjacency matrix, the MPF equivalent of Equation \ref{eq:GCN_architecture} is
\begin{equation}
    \vect{h}_i^{(l+1)} = \sigma \Bigg( \frac{1}{c_i} \Bigg[ \vect{h}_i^{(l)} + \sum_{j\in\mathcal{N}_i} \vect{h}_j^{(l)} \Bigg] \vect{W}^{(l)}\Bigg),
    \label{eq:GCN_message_passing}
\end{equation}
where $\vect{h}_i^{(l)}$ refers to the embedding of node $v_i$ in layer $l$, $\mathcal{N}_i$ denotes the set of direct neighbors of node $v_i$, and $c_i$ is a node-specific normalization constant, which typically takes a value of $|\mathcal{N}_i|+1$ in case of an undirected network. In case the graph has edge weights $w_{ij}$, this can be incorporated as follows:
\begin{equation}
    \vect{h}_i^{(l+1)} = \sigma \Bigg( \frac{1}{c_i} \Bigg[ w_s \vect{h}_i^{(l)} + \sum_{j\in\mathcal{N}_i} w_{ij} \vect{h}_j^{(l)} \Bigg ] \vect{W}^{(l)}\Bigg),
    \label{eq:GCN_message_passing2}
\end{equation}
with a normalization constant $c_i = w_s + \sum_{j\in\mathcal{N}_i} w_{ij}$.
Note that a self-weight $w_s$ has been introduced, which can also be turned into a trainable parameter.

\begin{figure*}
    \centering
    \includegraphics[width=\textwidth]{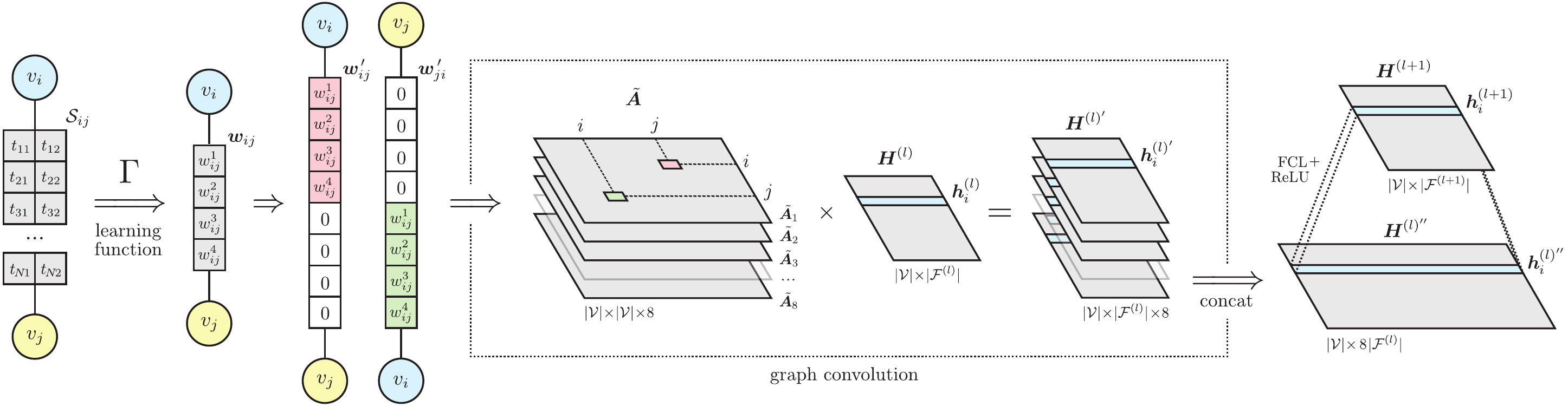}
    \caption{Schematic overview of a single L-GCN layer. A learning function $\Gamma$ transforms sets of edge attribute vectors $\mathcal{S}_{ij}$ into embeddings $\vect{w}_{ij} \in \mathbb{R}^L$
    , which are used to encode two similar but different latent relations in both canonical directions. These \emph{pseudo-relations} form a latent adjacency tensor, which is used in a graph convolution to generate a new set of node embeddings $\vect{H}^{(l+1)}$ (see Equation \ref{eq:RGCN_rewrite2}).}
    \label{fig:main_gcn}
\end{figure*}

\subsection{R-GCN}
\label{subsection:RGCN}
In knowledge graphs, each edge is annotated with a relation type, which can be represented by a one-hot encoded vector with a length equal to the cardinality of the set of all relations $\mathcal{R}$. \citeauthor{RGCN_paper} use these vectors to replace the adjacency matrix $\vect{A}$ from Equation \ref{eq:GCN_architecture} with an adjacency \emph{tensor} of shape $|\mathcal{V}|\times|\mathcal{V}|\times|\mathcal{R}|$, or as an MPF:
\begin{equation}
    \vect{h}_i^{(l+1)} = \sigma \Bigg( \vect{h}_i^{(l)}  \vect{W}_s^{(l)} + \sum_{r\in\mathcal{R}}\sum_{j\in\mathcal{N}_i^r} \frac{1}{c_{i,r}} \vect{h}_j^{(l)}  \vect{W}_r^{(l)} \Bigg),
    \label{eq:RGCN}
\end{equation}
where $\vect{W}_r \in \mathbb{R}^{|\mathcal{F}^{(l)}|\times|\mathcal{F}^{(l+1)}|}$ are \emph{relation-specific} weight matrices, $\vect{W}_s$ denotes the weight matrix associated with self-connections and $c_{i,r}$ are relation-specific normalization constants, typically of value $|\mathcal{N}_i^r|$. In order to facilitate the directional nature of some relations, $\mathcal{R}$ is expanded to contain the canonical as well as the inverse variations of each relation.

\subsection{Relation to Other Work}

Propagating edge features (beyond a finite set of edge labels) in graph models is not a new area of investigation. Common frameworks like neural message passing networks \citep{GilmerSRVD17} and graph networks \citep{battaglia2018relational} allow propagation of edge features: each layer contains learned edge features, derived from the previous layer. In our model, the information on the edges is aggregated once, and encoded into the node features and the edge \emph{weights}. 

While these frameworks allow for edge features through functions which propagate them, relatively little research exists on the question of how best to instantiate these functions. In \citep{kearnes2016molecular}, edge representations are explicitly included and updated through neural message passing in the context of molecular graphs. Such propagation is often framed as node-only message passing on the associated Levi graph \citep{levi1942finite,beck2018graph} (where both nodes and edges are turned into nodes in a bipartite graph). In our use cases, this approach would lead to a combinatorial explosion due to the large number of edges. We do, however, propagate from the edges to the nodes in the DVE mechanism described in Section~\ref{subsubsection:DVE}. 

In \cite{gong2019exploiting} and \cite{simonovsky2017dynamic}, edge features are exploited to learn the propagation weights, which is similar to the basic L-GCN presented here, yet without multiple (complex) edges. The principle of learning propagation weights is also used in the GAT model \citep{velivckovic2017graph}. In other work, e.g. \citep{battaglia2016interaction,chang2016compositional}, no edge features are provided in the input, but the models derive latent edge features that are useful in solving the downstream task.

What makes our use cases additionally unique are the multiple edges between nodes (up to thousands), whereas most research so far has focused on simple graphs or graphs with a fixed possible number of edges between two nodes, as in \citep{knyazev2018spectral,geng2019spatiotemporal,chai2018bike}. While the extension of message passing models to generic multigraphs with an unbounded number of edges between nodes has occasionally been mentioned (e.g. \cite[Section~2]{GilmerSRVD17}), to the best of our knowledge this is the first end-to-end model to tackle such a use case in practice.

Finally, it is important to note that our work can in fact be used to learn end-to-end from graphs with \emph{any} type of edge representation, whether this be text, images or vector time series, by picking a suitable learning mechanism.

\section{LATENT GRAPHS}
\subsection{L-GCN}
\label{subsection:LGCN}

To move from the explicit relations used in R-GCNs to \emph{latent} relations, we represent each edge $e_{ij} \in \mathcal{E}$ by L weights $w^r_{ij}$. The propagation rule takes a form akin to Equation \ref{eq:GCN_message_passing2}:
\small
\begin{equation}
    \vect{h}_i^{(l+1)} = \sigma  \Bigg( \frac{1}{c_i} \sum_{r\in\mathcal{R}} \Bigg[ w^r_s \vect{h}_i^{(l)} + \sum_{j\in\mathcal{N}_i} w^r_{ij} \vect{h}_j^{(l)} \Bigg] \vect{W}_r^{(l)} \Bigg),
    \label{eq:RGCN2}
\end{equation}
\normalsize
now with 
$c_i = \sum_{r\in\mathcal{R}}( w^r_s + \sum_{j\in\mathcal{N}_i} w^r_{ij})$.
$\vect{W}_s^{(l)}$ from Equation \ref{eq:RGCN} has been absorbed into $\vect{W}_r^{(l)}$; the weight of self-connections can now be learned by means of a self-weight vector $\vect{w}_s$. This method obtains an updated node representation for every \emph{pseudo-relation} separately, after which a weighted summation takes place to arrive at the final embedding\footnote{\citeauthor{RGCN_paper} allude to the possibility of replacing the weighted sum with something more elaborate, e.g. a fully connected layer, but left this for future work.}.


With \emph{Latent-Graph Convolutional Networks} (L-GCN), we turn the edge weights $w^r_{ij}$ from Equation \ref{eq:RGCN2} into trainable parameters in an end-to-end fashion by applying a learning mechanism to the information describing the edges, e.g. a sequence $\mathcal{S}_{ij}$ of  vectors:
\begin{equation}
    \vect{w}_{ij} = \Gamma(\mathcal{S}_{ij}),
    \label{eq:LGCN}
\end{equation}
where $\Gamma (\cdot)$ represents a differentiable learning function that is best suited for the data at hand. The information residing in the edge is embedded in a vector $\vect{w}_{ij} \in \mathbb{R}^L$ of length $L$, representing \emph{latent relations} between nodes $v_i$ and $v_j$. These weights then serve as input for the propagation rule from Equation \ref{eq:RGCN2}.

As with other encoding methods, the output layer of the learning function effectively serves as a bottleneck for the information that can flow from the edges to the vertices. The optimal size $L$ depends on both data and network complexity, and hence is best left to to hyperparameter tuning.

\paragraph{Bidirectionality}
We introduce propagation along both edge directions in a similar fashion as \citeauthor{RGCN_paper}, but adapted to Equation \ref{eq:RGCN2}. The actual embedding size becomes $2L$, with the original edges represented by (when $L=4$)
\begin{equation}
    \vect{w}_{ij}' = [w_{ij}^1, w_{ij}^2, w_{ij}^3, w_{ij}^4, 0, 0, 0, 0]^\intercal,
\end{equation}
and the inverse versions of the latent relations by 
\begin{equation}
    \vect{w}_{ji}' = [0, 0, 0, 0, w_{ij}^1, w_{ij}^2, w_{ij}^3, w_{ij}^4]^\intercal.
\end{equation}
The network can now process incoming and outgoing sets of edges in different, independent ways, allowing for a bidirectional propagation of information across the graph. Note that for a single L-GCN layer, $\Gamma (\cdot)$ needs to act on $\mathcal{S}_{ij}$ only once, as $\vect{w}_{ij}'$ and $\vect{w}_{ji}'$ use the same weights. Figure \ref{fig:main_gcn} shows a schematic overview of an L-GCN layer with bidirectional propagation.

\subsection{L-GCN+}
\label{subsection:LGCN_plus}

Equation \ref{eq:RGCN2} can be reformulated to first aggregate over all pseudo-relations $\mathcal{R}$ and then aggregate over all neighbours $\mathcal{N}_i$ as follows:
\scriptsize
\begin{equation}
    \vect{h}_i^{(l+1)} = \sigma \Bigg(\frac{1}{c_i} \sum_{r\in\mathcal{R}} w^r_s \vect{h}_i^{(l)} \vect{W}_r^{(l)} + \frac{1}{c_i} \sum_{j\in\mathcal{N}_i} \sum_{r\in\mathcal{R}} w^r_{ij} \vect{h}_j^{(l)} \vect{W}_r^{(l)} \Bigg).
    \label{eq:RGCN_rewrite}
\end{equation}
\normalsize
We can now also replace all relation-specific weight matrices $\vect{W}_r \in \mathbb{R}^{|\mathcal{F}^{(l)}|\times|\mathcal{F}^{(l+1)}|}$ by a single weight matrix $\vect{W} \in \mathbb{R}^{L\cdot|\mathcal{F}^{(l)}|\times|\mathcal{F}^{(l+1)}|}$, representing a fully connected layer:
\scriptsize
\begin{equation}
    \vect{h}_i^{(l+1)} = \sigma \Bigg( \frac{1}{c_i} \vect{w}_s \otimes \vect{h}_i^{(l)} \vect{W}^{(l)}  + \sum_{j\in\mathcal{N}_i} \frac{1}{c_i} \vect{w}_{ij} \otimes \vect{h}_j^{(l)}\vect{W}^{(l)} \Bigg).
    \label{eq:RGCN_rewrite2}
\end{equation}
\normalsize
This enables a straightforward introduction of nonlinear interactions between edge embeddings and node attributes on a per-neighbor basis, by introducing an intermediate layer of size $2\cdot |\mathcal{F}^{(l+1)}|$, splitting the original transformation into two matrices $\vect{W}_1 \in \mathbb{R}^{L\cdot|\mathcal{F}^{(l)}|\times2\cdot|\mathcal{F}^{(l+1)}|}$ and $\vect{W}_2 \in \mathbb{R}^{2\cdot|\mathcal{F}^{(l+1)}|\times|\mathcal{F}^{(l+1)}|}$:
\footnotesize
\begin{equation}
    \vect{h}_i^{(l+1)} = \sigma \Bigg( f \Big( \frac{1}{c_i} \vect{w}_s \otimes \vect{h}_i^{(l)} \Big) + \sum_{j\in\mathcal{N}_i} f \Big( \frac{1}{c_i} \vect{w}_{ij} \otimes \vect{h}_j^{(l)} \Big) \Bigg),
    \label{eq:RGCN_plus}
\end{equation}
\normalsize
with $f(\cdot)$ a simple, two-layer MLP with its own nonlinearity:
\begin{equation}
    f^{(l)}(\vect{x}) = \sigma \Big( \vect{x}\vect{W}^{(l)}_1 \Big) \vect{W}^{(l)}_2.
    \label{eq:RGCN_plus2}
\end{equation}
We conjecture that this modification enables the network to learn more complex patterns \emph{within} the interactions between node features and the latent-relation attributes\footnote{For sake of brevity, we have omitted biases from all equations. In our implementations, weight matrices $\vect{W}\in \mathbb{R}^{A\times B}$ are accompanied by biases $\vect{b} \in \mathbb{R}^{B}$}.

\section{METHODS}

\subsection{Data Sets}

\subsubsection{Financial Transactions}
The synthetic data sets we use in our experiments simulate fraud in financial transaction networks. 42k vertices $v_i \in \mathcal{V}$ represent actors in the network, and (directed) multi-edges $e_{ij} \in \mathcal{E}$ represent 125k sequences of transactions $\mathcal{S}_{ij}$ from $v_i$ to $v_j$ ($\sim$6.5M transactions in total). Labels are only provided at the node level ($C=2$ classes: fraud \textbf{F} and normal \textbf{N} with ratio 1:9), but relevant patterns are hidden in the transaction sets.

All nodes have a set of attributes $\mathcal{F}$, i.i.d. sampled from same distributions, irrespective of the node class (and therefore cannot be used for node classification inductively). All edges represent sets of transaction vectors $\vect{t}_k \in \mathcal{S}_{ij}$, belonging to one of three types (weekly transactions, monthly, or randomized w.r.t time). In case \emph{one} of the associated nodes belongs to \textbf{F} (fraud), mutations are applied to the sets depending on the direction (i.e. double or missing transaction). It is \emph{these} patterns the network will need to learn to perform the downstream task.

\begin{figure}[t]
    \centering
    \includegraphics[width=\columnwidth]{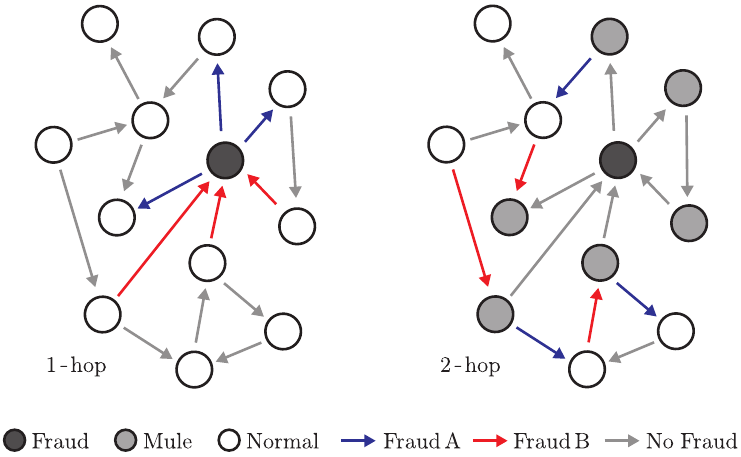}
    \caption{Schematic examples of a 1-hop structure (left) versus a 2-hop structure (right). In the 2-hop structure, fraudulent activity takes place at least once removed from the fraudulent actor in the network.}
    \label{fig:1hop2hop}
\end{figure}

To facilitate an additional challenge and an interesting comparison, we also generate a second synthetic data set, in which we introduce these mutations at a 2-hop distance from the fraud nodes. An illustration of this difference in structure can be seen in Figure \ref{fig:1hop2hop}. Relevant information is now hidden deeper into the graph.

Transactions between the same vertices but in opposite direction are treated as two distinct edge populations $\mathcal{S}_{ij}$ and $\mathcal{S}_{ji}$. Individual transactions come with two properties $t_{k1}$ and $t_{k2}$ (time delta $\Delta t$ and amount). For a more detailed explanation of these data sets we refer to the supplementary material provided.

\begin{table}[t]
    \caption{Overview of data set characteristics: number of vertices, multi-edges and total edges.}
    \begin{tabularx}{\columnwidth}{X c c c}
         \textbf{Data Set} & $|\mathcal{V}|$ & $|\mathcal{E}|$ & $\sum_\mathcal{E}|\mathcal{S}|$\\
        \hline
        \hline
        \textbf{Financial} (1-hop) & 41792 & 124996 & 6643964\\
        \textbf{Financial} (2-hop) & 41758 & 124995 & 6618483\\
        \textbf{Transportation} & 2940 & 25000 & 4200000\\
        \hline
    \end{tabularx}
    \label{table:datasets}
\end{table}

\subsubsection{Urban Transportation} The real-world scenario we use to demonstrate our architecture represents a combination of data from three different sources, all related to the New York City borough of Manhattan. We use data from the 2010 US decennial census to retrieve 6 data points at the census-block level (roughly corresponding to regular city blocks), related to total population, median age and housing unit occupation / ownership \citep{2010census}. These blocks represent our nodes with their associated attributes.

Next, we use official 2012\footnote{2012 is the earliest year in which zoning lots are mapped to 2010 census blocks} zoning data \citep{2012zoning} in order to determine the zoning type most prevalent in each census block, which can now all be assigned one of $C=3$ classes: residential \textbf{R}, commercial \textbf{C} and manufacturing \textbf{M}, with ratios $\sim$5:3:1. Predicting these classes will be our downstream task. The combination of these two data sets leaves us with 2940 usable nodes.

Finally, we look at all $\sim$160M Yellow Taxi rides made in 2012 and their associated tips \citep{2012taxi}, by mapping pickup and drop-off coordinates to the census block map. We then sample the 25k most popular routes to get a directed graph in which sets of taxi rides represent our edges. Due to memory constraints, it is not feasible to keep the taxi rides in their raw form. We therefore aggregate all rides by day of the week and hour of the day, each edge being described by a matrix $\vect{S}_{ij} \in \mathbb{R}^{168\times2}$. We conjecture that characteristics of these activity and tip profiles can be used to enhance zoning type prediction capability in an end-to-end fashion . For a more detailed explanation of this data set we refer to the supplementary material.

The slight mismatch in years between the sources, potential inaccuracy of taxi coordinate data and reduction of every census block to a single zoning type will inherently limit the obtainable accuracy by our models. We nevertheless hypothesise that the taxi network can serve as an added benefit.

\subsection{Architecture Components}
\label{subsection:components}

\subsubsection{Learning Mechanisms}
\label{subsubsection:learningmechanisms}
\begin{figure}[t]
    \centering
    \includegraphics[width=\columnwidth]{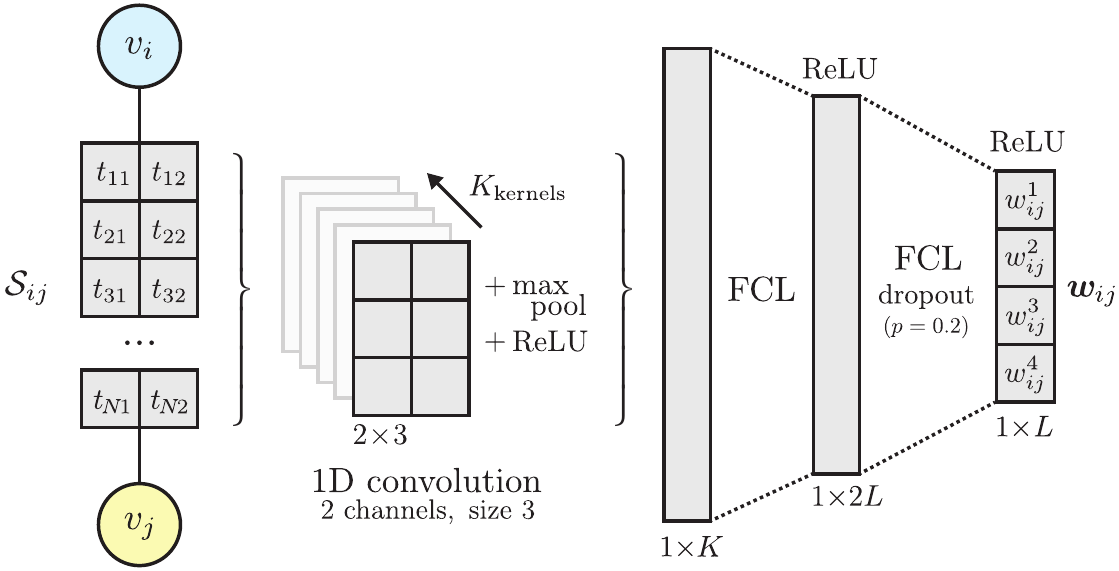}
    \vspace{0.5cm}
    \caption{$\Gamma$ ($L=4$) for the synthetic financial transaction data sets. Sequences of transactions between vertices are transformed into a single vector representation, embedding the latent relation.}
    \label{fig:MEE}
\end{figure}

\paragraph{Financial Transactions}
The learning mechanism $\Gamma$ (see Figure \ref{fig:MEE}) consists of a 1D convolution with $K$ kernels (size 3,  stride 1) and $Z$ channels, that slide over the sequences of transactions $\mathcal{S}_{ij}$. $Z$ corresponds with the number of attributes for each transaction, which is 2 in this case ($t_{k1}$ and $t_{k2}$). We perform a 1D max pooling operation across the output of each kernel (of shape $K \times (|\mathcal{S}_{ij}|-2)$) followed by an activation function $\text{ReLU}(\cdot)=\text{max}(0,\cdot)$. The resulting layer of size $K$ is  followed by two fully connected layers of sizes $2L$ and $L$ to obtain the desired embedding size. Activation functions (also ReLU) are employed after every fully connected layer, and dropout ($p=0.2$) is applied to the second-to-last layer (of size $2L$).

\paragraph{Transportation Network}
We aggregate the route (edge) profiles into a representation of the average 24-hour cycle with 2-hour bins (size $12\times2$). Due to strong Pareto-distribution effects, most of the 25k edge profiles are not based on enough individual rides to use more granular representations. The resulting layer of size 24 is followed by two hidden layers of size $2L$ and $L$, each accompanied by sigmoid activations.

\subsubsection{DVE (Direct on-Vertex Embedding)}
\label{subsubsection:DVE}
\begin{figure}[t]
    \centering
    \includegraphics[width=0.95\columnwidth]{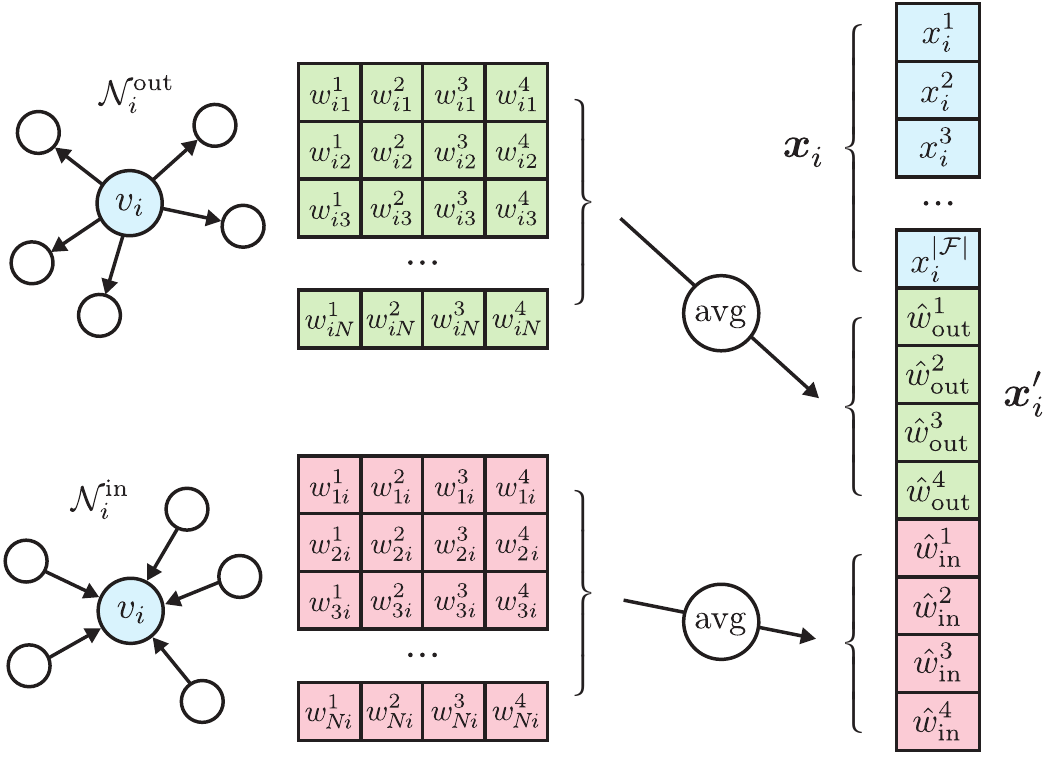}
    \vspace{0.5cm}
    \caption{Schematic overview of the DVE mechanism. A $\Gamma$ learning function ($L=4$) is applied to edges connecting $v_i$ to local neighborhoods $\mathcal{N}_i^{\mathrm{in}}$ and $\mathcal{N}_i^{\mathrm{out}}$ separately. The results are averaged and used to expand the original node features from $x_i$ to $x_i'$.}
    \label{fig:DVE}
\end{figure}

The objective of the learning function $\Gamma$ is to produce a vector $\vect{w}_{ij}$ that embeds data residing on edges in $\mathbb{R}^L$. Further propagation does not necessarily require integration into a GCN, one could, for instance resort to a simple local neighborhood aggregation. To serve as an additional baseline, we therefore introduce the \emph{Direct on-Vertex Embedding} mechanism, in which $\Gamma$ operates on each multi-edge population $\mathcal{S}_{ij}$ that connect a vertex $v_i$ to its local neighborhood $\mathcal{N}_i$ (see Figure \ref{fig:DVE}). We average the outputs of the learning mechanism and append the result to the original node features. This way, information related to vertices residing in their associated edges is directly incorporated in updated node embeddings. We expand the vertex embedding twice, for incoming and outgoing edge populations separately, to allow for bidirectional propagation and different treatment of each canonical direction.

\subsection{Overview of the Architectures}
\label{subsection:architectures}

\paragraph{GCN}
The standard GCN architecture (Equation \ref{eq:GCN_message_passing}) will be used as a baseline. To allow for a two-way flow of information across the graph, the original (directed) adjacency matrices are transformed into their undirected counterparts via $\vect{A}'=\vect{A}+\vect{A}^T$. Note that an edge now only indicates there \emph{exist} a connection between the two nodes. More detailed information is not taken into account.

\paragraph{DVE}
A non-GCN baseline architecture based on a single local neighborhood aggregation using the DVE mechanism. After the expansion of the node features, two fully connected layers follow, with sizes corresponding to the intermediate layers of the upcoming L-GCN architecture layers.

\paragraph{LX-GCN}
Architectures in which the $\Gamma$ function provides embedding vectors $\vect{w}_{ij} \in \mathbb{R}^L$ with $L=X$. Further propagation takes place according to the rule from Equation \ref{eq:RGCN_rewrite2}. A schematic overview of the $L=4$ variation can be seen in Figure \ref{fig:main_gcn}.

\paragraph{LX-GCN+}
Similar to LX-GCN, with additional nonlinear interaction on a per-neighbor basis, as described in Section \ref{subsection:LGCN_plus}. Propagation takes place according to the rule from Equation \ref{eq:RGCN_plus}.

\begin{table*}
    \fontsize{8}{9}\selectfont
    \caption{Model performance on the transaction data sets. Accuracy and AUC (area under the ROC curve) are averaged over 10 runs and accompanied by their standard error. Also displayed are the duration of the training sessions and required GPU RAM allocation. Training was performed on Nvidia V100 Tensor Core GPUs.}
    \begin{tabularx}{\textwidth}{X | l@{ }l l@{ }l | l@{ }l l@{ }l | c c c }
         & \multicolumn{4}{c|}{\textbf{1-Hop}} & \multicolumn{4}{c|}{\textbf{2-Hop}} & & & \\
        \textbf{Architecture} & \multicolumn{2}{c}{\textbf{Accuracy}} & \multicolumn{2}{c|}{\textbf{AUC}} & \multicolumn{2}{c}{\textbf{Accuracy}} & \multicolumn{2}{c|}{\textbf{AUC}} & \textbf{Time} & \textbf{GPU RAM} & 
        $\boldsymbol{N}_{\textbf{params}}$ \\
        \hline
        \hline
        \multicolumn{12}{c}{Baselines} \\
        \hline
        \textbf{Random Guess}    & 50.06 & $\pm\ 0.08$ & 0.500 & $\pm\ 0.002$ & 49.90 & $\pm\ 0.09$ & 0.497 & $\pm\ 0.001$  &               &          & \\
        \textbf{Majority Class}  & 89.91 &             & 0.500 &              & \textbf{90.01} &             & 0.500 &               &               &          & \\
        \textbf{GCN}             & 49.62 & $\pm\ 0.27$ & 0.501 & $\pm\ 0.001$ & 56.50 & $\pm\ 0.44$ & 0.513 & $\pm\ 0.001$  & $\sim 17$s    & 1.82 GB  & 322 \\
        \textbf{DVE} ($L=4$)     & \textbf{97.41} & $\mathbf{\pm\ 0.15}$ & 0.969 & $\pm\ 0.007$ & 80.22 & $\pm\ 0.74$ & 0.889 & $\pm\ 0.012$  & $\sim 5$m     & 4.64 GB  & 746 \\
        \hline
        \multicolumn{12}{c}{Prototypes} \\
        \hline
        \textbf{L1-GCN}          & 56.49 & $\pm\ 2.83$ & 0.514 & $\pm\ 0.013$ & 61.85 & $\pm\ 1.76$ & 0.556 & $\pm\ 0.028$  & $\sim 9$m     & 7.58 GB & 994 \\
        \textbf{L2-GCN}          & 86.94 & $\pm\ 5.29$ & 0.862 & $\pm\ 0.061$ & 80.13 & $\pm\ 2.41$ & 0.862 & $\pm\ 0.045$  & $\sim 10$m    & 7.96 GB & 1694 \\
        \textbf{L2-GCN+}         & 95.38 & $\pm\ 1.13$ & 0.970 & $\pm\ 0.006$ & 85.50 & $\pm\ 1.46$ & 0.917 & $\pm\ 0.034$  & $\sim 10$m    & 8.11 GB & 3746 \\
        \textbf{L4-GCN}          & \textbf{97.10} & $\mathbf{\pm\ 0.29}$ & 0.972 & $\pm\ 0.004$ & 89.65 & $\pm\ 0.34$ & \textbf{0.952} & $\mathbf{\pm\ 0.004}$  & $\sim 11$m    & 8.52 GB & 3118 \\
        \textbf{L4-GCN+}         & \textbf{97.30} & $\mathbf{\pm\ 0.41}$ & \textbf{0.983} & $\mathbf{\pm\ 0.002}$ & 87.34 & $\pm\ 1.26$ & \textbf{0.951} & $\mathbf{\pm\ 0.004}$  & $\sim 11$m    & 8.47 GB & 6370 \\
        \hline
    \end{tabularx}
    \label{table:results}
\end{table*}

\subsection{Training \& Hyperparameters}
\label{subsection:hyperparameters}
We use a weighted version of cross-entropy loss for all of our architectures and data sets, as is appropriate for classification tasks with imbalanced classes. All training sessions employ the Adam optimizer.

For both synthetic data sets we use a fixed training/validation/test split scheme with ratios 5/5/90, as is standard practice for semi-supervised learning on graphs. For the urban transportation data set we used ratios 60/20/20 because there are significantly fewer nodes. Validation sets were used during architecture and hyperparameter optimization. Test sets were held out entirely, until final settings and architectures were decided on. 

All (L-)GCN architectures consist of two (L-)GCN layers. As common with GCN-like architectures, adding more layers does not provide any noticeable benefit, which may be related to excessive information diffusion. In all transaction network architectures, we apply dropout ($p=0.5$) to the output of the \emph{first} GCN-layer for regularization purposes. Note that each of the two GCN layers of every L-GCN-like architecture contain their \emph{own} instance of the $\Gamma$ learning function, allowing for different embeddings to be learned in each stage of the propagation.

Hyperparameter sweeps were performed to find good node embedding sizes for $\vect{H}^{(1)}$ at layer 1, learning rate and weight decay. In case of the transaction networks, a number of convolutional kernels $K=20$ for $\Gamma$ was found to be optimal with respect to other options explored, together with a learning rate of $5\cdot 10^{-4}$, weight decay of $5\cdot 10^{-4}$ and an intermediate embedding size of 20 for $\vect{H}^{(1)}$. We run all training sessions for 2000 epochs, after which training settles into an equilibrium. For the urban transportation case, a learning rate of $1.5\cdot 10^{-4}$, weight decay of $5\cdot 10^{-4}$, an intermediate embedding size of 6 for $\vect{H}^{(1)}$ was used and all models were trained for 2000 epochs. Implementation details can be found in the supplementary material provided.

\section{RESULTS}
\subsection{Transaction Networks}
All models were subjected to 10 training sessions with independent, random weight initialization and the settings described in Section \ref{subsection:hyperparameters}. The resulting accuracy scores on the test sets are averaged and displayed in Table \ref{table:results}, accompanied by their standard error. Note that the resulting variance is only due to a different initialization of weights, as the training/validation/test splits do not vary. Because of the severe class imbalance, accuracy scores tend to be misleading. For the transaction networks, we therefore also display the AUC (area under the receiver operating characteristic curve), obtained from the original, decimal values of the model output \citep{SPACKMAN1989160}.

\subsubsection{1-Hop Data Set}
Since node attributes $\mathcal{F}$ are i.i.d. sampled for both classes, the classic GCN architecture is not able to improve on random guessing. This is to be expected as there is no access to transaction-level information.

The L1-GCN architecture only performs slightly better than the GCN, which should be attributed to the fact that it can only use coincidental statistical fluctuations in the node attributes, as it only assigns a learned degree of weight to neighbors. As we expand the size of the latent representation, we see significant increased performance and a clear advantage of $L=4$ over $L=2$. Other latent representation sizes $L$ were not found to increase performance.

The additional introduction of the nonlinear interactions on a per-neighbor basis (L-GCN+) results in marked improvement over the ``naive" graph convolutional counterparts, even bringing L2-GCN+ close to L4-GCN. The best L-GCN architectures perform on par with the DVE architecture. We are reluctant to draw any further conclusions since model performances approach 100\% accuracy. We also conjecture that L4-GCN and L4-GCN+ perform similarly because we approach an information limit\footnote{We do not perform any additional statistical tests on the performance samples since they are all produced by the same training/validation/test splits and hence not independent.}.

\subsubsection{2-Hop Data Set}
Whereas the DVE baseline architecture performed well on the 1-hop data set, its AUC and accuracy score are now significantly lower. This is in line with expectations, since a lot of relevant information can no longer be found within the local neighborhood. The performance of the L-GCN prototypes still increases with each added level of complexity in a similar fashion, albeit at slightly lower performances (both accuracy and AUC). The 2-hop structure must therefore pose a more difficult classification challenge to our models, as relevant information is hidden at a deeper level in the data set. The L4-GCN(+) architectures, however, still perform well with accuracy scores close to 90\% and AUC scores over 0.95.

\subsubsection{Inductive Setting}
Our assessment so far has taken place in a \emph{transductive} setting: inference is performed on the same graph seen during training. This is common in the realm of GCNs and covers many real-world scenarios. It is equally relevant, however, to investigate how well our trained models generalize to different graphs of the same type, or a later versions of the same graphs. An obvious use case in this \emph{inductive} setting is graphs subject to time evolution, for which generalization could eliminate the need for frequent retraining. We generate different versions from the same distributions of both the 1-hop and 2-hop data sets (new graph structures, transaction sets \emph{and} node attributes) and apply the same instances of our trained architectures.

Looking at the results in Table \ref{table:inductive}, the first thing to notice is that all models show much lower levels of performance. Remarkably, the DVE baseline architecture (non-GCN) now even performs on par with random guessing, indicating that all patterns learned are specific to the graph used during training and do not generalize \emph{at all}. The L1-GCN architecture now also performs similar to random guessing, which confirms the conjecture that the little performance shown in the transductive setting is indeed related to relying on graph-specific statistical fluctuations in the node attributes.

Other L-GCN architectures do generalize to an extent, however, L4-GCN+ even obtaining an average AUC score over 0.90 on the 1-hop data set. Allowing for nonlinear interaction \emph{before} the local neighborhood aggregation seems to be crucial, with those versions outperforming their normal counterparts by a wide margin. This is one of the more interesting findings of this study, as the L-GCN+ architectures are more complex than their normal variations, containing \emph{over twice} as many model parameters (see Table \ref{table:results}). This makes the fact that they are more inclined to generalize to unseen networks somewhat counter-intuitive. We conjecture that allowing nonlinear interactions on a per-neighbor basis favors the learning of generalized patterns but leave a more in-depth investigation of this phenomenon to future work, including to which extent this behaviour can be replicated for other GCN-like architectures.

\begin{table}
    \fontsize{8}{9}\selectfont
    \caption{Model performance (AUC) in an inductive setting. AUC values are averaged over the same 10 runs from Table \ref{table:results} and accompanied by their standard error.}
    \begin{tabularx}{\columnwidth}{X | l@{ }l | l@{ }l }
        \textbf{Architecture} & \multicolumn{2}{c|}{\textbf{1-Hop AUC}} & \multicolumn{2}{c}{\textbf{2-Hop AUC}} \\
        \hline
        \hline
        \multicolumn{5}{c}{Baselines} \\
        \hline
        \textbf{GCN} & 0.500 & $\pm\ 0.000$ & 0.500 & $\pm\ 0.001$  \\ 
        \textbf{DVE} ($L=4$) & 0.499 & $\pm\ 0.001$ & 0.500 & $\pm\ 0.000$  \\ 
        \hline
        \multicolumn{5}{c}{Prototypes} \\
        \hline
        \textbf{L1-GCN} & 0.500 & $\pm\ 0.000$ & 0.500 & $\pm\ 0.000$  \\ 
        \textbf{L2-GCN} & 0.581 & $\pm\ 0.041$ & 0.639 & $\pm\ 0.046$  \\ 
        \textbf{L2-GCN+} & 0.859 & $\pm\ 0.055$ & \textbf{0.816} & $\mathbf{\pm\ 0.041}$  \\ 
        \textbf{L4-GCN} & 0.653 & $\pm\ 0.051$ & 0.575 & $\pm\ 0.024$  \\ 
        \textbf{L4-GCN+} & \textbf{0.907} & $\mathbf{\pm\ 0.051}$ & 0.784 & $\pm\ 0.049$  \\
        \hline
    \end{tabularx}
    \label{table:inductive}
\end{table}

\subsection{Transportation Network}
We evaluate a similar range of models on the urban transportation data set, accompanied by some typical multinomial classification methods 
(given a full hyperparameter sweep). 
Since the classes are again imbalanced, we also display the macro-F1 score along with the accuracy (see Table \ref{table:inductive}).

We find that the taxi ride network indeed contains useful information w.r.t. zoning category prediction. Baseline methods that do not have access to the network structure have low predictive power. 
A simple GCN already outperforms these significantly, showing that providing access to just the network structure already boosts performance.

Models with end-to-end access to edge-level information however, show still higher performance. And even though the DVE and L-GCN architectures perform well, they are once again outperformed by L-GCN+ (other latent representation sizes $L$ were not found to increase performance). The findings in this real-world scenario, although in a different application domain, are largely in line with our earlier observations.

As this concerns a novel problem domain without existing end-to-end solutions, we by no means claim that these added baselines are the best possible alternatives; they are there to provide context for our results. We invite the community to experiment with our data sets and explore alternative approaches.

\begin{table}[t]
    \caption{Model performance (accuracy and macro-F1) on the urban transportation data set. Scores are averaged over 10 runs and accompanied by their standard error.}
    \fontsize{8}{9}\selectfont
    \begin{tabularx}{\columnwidth}{X | l@{ }l | l@{ }l }
        \textbf{Architecture} & \multicolumn{2}{c|}{\textbf{Acccuracy}} & \multicolumn{2}{c}{\textbf{Macro-F1}} \\
        \hline
        \hline
        \multicolumn{5}{c}{Baselines} \\
        \hline
        \textbf{Log. Regression} & 0.583 &  & 0.456 &  \\
        \textbf{Decision Tree} & 0.546 &  & 0.422 & \\
        \textbf{k-NN} & 0.680 & & 0.460 & \\
        \textbf{GCN} & 0.652 & $\pm\ 0.009$ & 0.549 & $\pm\ 0.011$  \\
        \textbf{DVE} ($L=3$) & 0.682 & $\pm\ 0.010$ & 0.593 & $\pm\ 0.009$  \\
        \hline
        \multicolumn{5}{c}{Prototypes} \\
        \hline
        \textbf{L3-GCN} & 0.678 & $\pm\ 0.009$ & 0.595 & $\pm\ 0.011$  \\ 
        \textbf{L3-GCN+} & \textbf{0.715} & $\mathbf{\pm\ 0.019}$ & \textbf{0.637} & $\mathbf{\pm\ 0.020}$  \\
        \hline
    \end{tabularx}
    \label{table:transportation}
\end{table}

\section{DISCUSSION \& CONCLUSIONS}

Classical GCNs essentially are information diffusion mechanisms through which data residing in node attributes can be smoothed over local graph neighborhoods. Because this diffusion is performed \emph{indiscriminately}, their success relies on the existence of an edge $e_{ij}$ being correlated with vertices $v_i$ and $v_j$ belonging to the same class. \citeauthor{KipfW16} showed that on data sets that fit that description, such as Citeseer, Cora and Pubmed, GCNs were indeed able to improve on state-of-the-art node classification results. When networks do not exhibit this correlation however, a GCN is not likely to provide better classification accuracy than random guessing. This creates a need for architectures that exploit more forms of information residing in networks as well as propagate information along the graph in a \emph{discriminative} fashion.

In this study, we have shown that we can perform end-to-end learning on complex multigraphs in two distinctively different settings, proposing Latent-Graph Convolutional Networks (L-GCNs). A suitable learning mechanism embeds multi-edges into latent relations, serving as input for GCN-like further propagation in the form of a latent adjacency tensor.

We have shown that our prototypes perform well on two synthetic data sets with relevant information hidden at different depths inside the network structure. Especially the data set with 2-hop correlations, our best models significantly outperform the non-GCN baseline architecture. We have also demonstrated L-GCNs in a real-world case in the form of an urban transportation network, in which the relative model performances match our earlier observations. Note that our approach can be used to learn end-to-end from \emph{any} type of edge representation, e.g. text or images, by picking a suitable learning mechanism.

The optimal number of latent relations $L$ depends on both data and network complexity, but we can conclude that $L=1$ does not offer enough degrees of freedom for the effective integration of information residing in the edges into the node embeddings, as it can only indicate the relative importance that needs to be assigned to a neighboring node. 

Introducing nonlinear interactions on a per-neighbor basis (i.e. \emph{before} local neighborhood aggregation) with L-GCN+ generally improves our models, but seems to be particularly beneficial in an inductive setting. This makes (L-)GCN+ architectures potentially very useful, e.g. for graphs subject to time evolution, and calls into question standard GCN-like approaches in inductive settings, given the performance drop. 
We leave further investigation of this to future work.

\bibliography{references.bib}

\appendix

\section{IMPLEMENTATION DETAILS}
All architectures are built on top of PyTorch Geometric, an extension library for PyTorch centered around GCN-like architectures \citep{Fey/Lenssen/2019}. Training is performed on Nvidia V100 Tensor Core GPUs.

All edge attributes are stored in a single tensor, with multi-edge sequences smaller than $|\mathcal{S}_{ij}|_{\textrm{max}}$ padded with zeros. This is relevant for our financial transactions case but does not significantly affect training due to the nature of $\Gamma$. It does however severely impact memory allocation on the GPU. We therefore choose to facilitate the (optional) grouping and batching of edge populations based on their number of edges $|\mathcal{S}_{ij}|$. The net effect of this batching is a memory reduction of $\sim$6$\times$ and a training speed increase of $\sim$2$\times$. Note that training still takes place in a full-batch fashion; this grouping is solely employed to reduce the amount of padding.

\section{TRANSPORTATION NETWORK DETAILS}
The real-world network we use to demonstrate how L-GCNs can be applied in multiple settings centers around 160M New York City Yellow Taxi rides made in 2012 \citep{2012taxi}. If we map all rides to discrete locations they form a multigraph, with these locations as vertices. To define these locations, we turn to the 2010 US decennial census \citep{2010census}, which specifies demographics at the census-block level (roughly corresponding to regular city blocks). For our target variables, we turn to the zoning information as published by the city council \citep{2012zoning}. We can use this data to find out for every census block which zoning type is most prevalent.

\subsection{Node Attributes}
For every census block in the New York City borough of Manhattan, we extract the following data attributes from the 2010 decennial census:
\begin{enumerate}
    \itemsep0em
    \item Population
    \item Median Age
    \item Occupied Housing Units
    \item Renter Owned
    \item Mortgage Owned
    \item Owned Free
\end{enumerate}
There were more data points available but we deemed these unfit for the current research. We applied the following normalization procedure: data point 2 was expressed as a fraction of the global maximum, points 1 and 3 were transformed into their log values and then also expressed as a fraction of the global maximum, while data points 4 through 6 were expressed as fractions of data point 3.

\begin{figure}[t]
    \centering
    \includegraphics[width=\columnwidth]{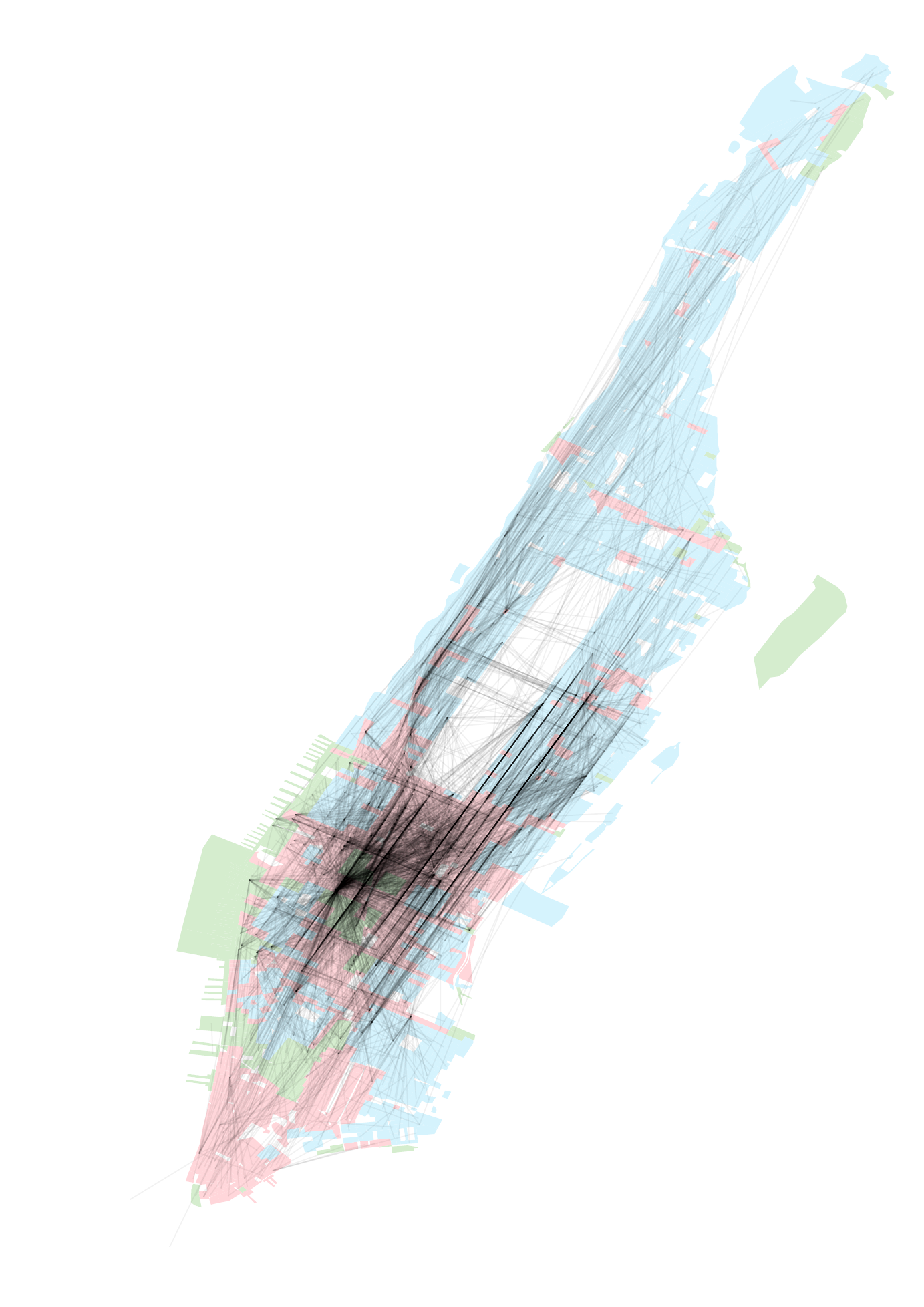}
    \caption{2012 Manhattan Yellow Taxi ride network between census blocks as given by the 25k most popular block-to-block combinations, with the exception that each block is represented at least twice (a random 20\% subset of edges is shown). Colors represent most prevalent zoning type for each block in 2012: residential (blue), commercial (pink) or manufacturing (green).}
    \label{fig:network_map}
\end{figure}

\subsection{Target Variables}
Zoning information for New York City comes in a the form of a collection of zoning lots, of which there are one or several for each census block. For each census block, we sum the areas for each zoning type (residential, commercial or manufacturing\footnote{We ignore parks and special zones such as Battery Park.}). We assign each census block the zoning type that covers the most area of that block. This is a simplified representation of reality but does not undermine the central part of the demonstration. The resulting ratios between residential / commercial / manufacturing are close to 5 / 3 / 1. The resulting map of census blocks and their associated most prevalent zoning type can be seen in Figure \ref{fig:network_map}.

\subsection{Network Structure}
In order to assign each of the $\sim$160M yellow taxi rides to a pair of vertices in our graph, we must map their pickup and drop-off locations to a census block. Fortunately, coordinates for these these locations are provided in the 2012 data set. Comparing all coordinates to the polygon shapes is computationally expensive, so we compute the Euclidean distances to all census block centroids, which we pre-compute from the coordinate polygon shapes (see Figure \ref{fig:network_map}). The shortest distance is taken to be the matching census block, provided it's below a certain threshold in order to discard rides entering and leaving Manhattan. Next, we extract the time and tip amount from each ride, which together make up a vector representation of a single edge in our multigraph.

The resulting $\sim$140M valid rides are too many to keep in raw form with respect to GPU RAM allocation during model training. We therefore choose to aggregate the rides by day of the week and hour of day, and collect the associated tips in a carbon copy of this array of bins. We perform this collection of data in two passes in order to preserve RAM.

During the first pass, for all of the possible block-to-block combinations ($\sim$10M), the number of rides is aggregated. We then select the top 25k routes in terms of total rides, provided that each census block is represented \emph{at least} twice\footnote{GPU RAM also places a limit on the number of edges that can effectively be handled during training. Also beyond 25k edges it is doubtful whether there are enough statistics to provide an added benefit during training since taxi rides unsurprisingly turn out to be Pareto distributed over the potential routes.} (one incoming and one outgoing). The resulting graph, mapped to the same census blocks can be seen in Figure \ref{fig:network_map}. During the second pass, we collect the rides and their associated tips for the aforementioned 25k selected routes and sort this information by day of the week and hour of the day in a matrix $\vect{S}_{ij} \in \mathbb{R}^{168\times2}$.

\begin{figure}[t]
    \centering
    \includegraphics[width=\columnwidth]{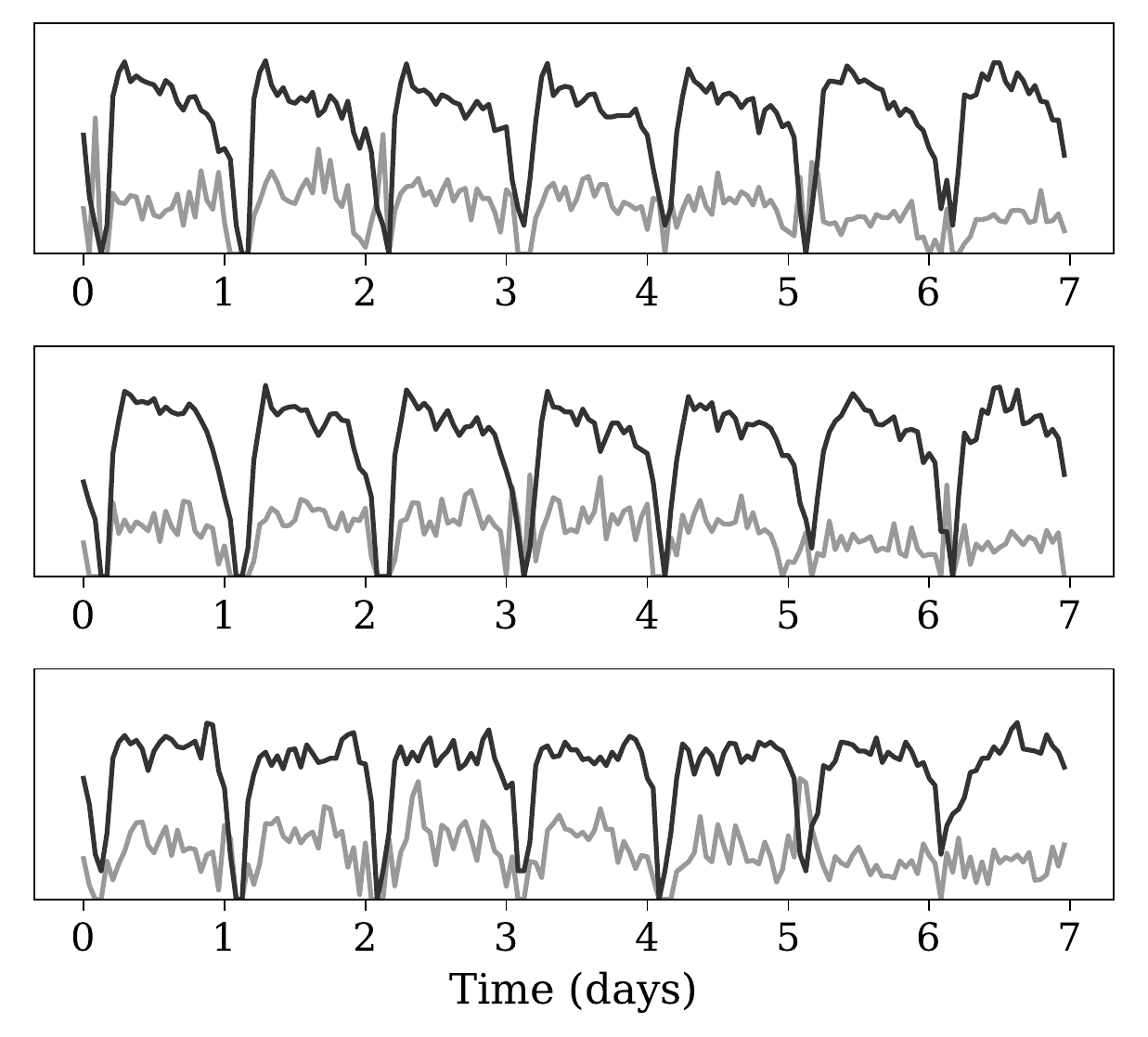}
    \caption{Activity (black) and average tip (gray) profiles for the three most-travelled routes in the 2012 Yellow Taxi / census block network. Ride characteristics are aggregated by day of the week and hour of the day in a matrix $\vect{S}_{ij} \in \mathbb{R}^{168\times2}$.}
    \label{fig:edge_profiles}
\end{figure}

For normalization purposes, all ride counts as well as average tip in each time bin are transformed into their log values and expressed as a fraction of their respective \emph{global} maximum. This way we preserve both local \emph{and} global structure in our data\footnote{Our data sets and implementations can be found on GitHub: github.com/florishermsen/L-GCN}. Three examples of the resulting profiles that characterize each edge can be seen in Figure \ref{fig:edge_profiles}. An overview of data set statistics can be seen in Table \ref{table:datasets_transportation}.

\begin{table}[t]
    \captionof{table}{Characteristics of the urban transportation network: number of vertices and edges, sizes of classes \textbf{R}, \textbf{C} and \textbf{M}, and total number of edge vectors.}
    \label{table:datasets_transportation}
    \vspace{0.1cm}
    \begin{tabularx}{\columnwidth}{X X X X X c}
         $|\mathcal{V}|$ & $|\mathcal{E}|$ & $|\mathbf{R}|$ & $|\mathbf{C}|$ &  $|\mathbf{M}|$ & $\sum_\mathcal{E}|\mathcal{S}|$\\
        \hline
        \hline
        2940 & 25000 & 1671 & 1011 & 258 & 4200000\\
        \hline
    \end{tabularx}
\end{table}

\section{EMBEDDING INSPECTION}
\begin{figure}[t]
    \centering
    \includegraphics[width=\columnwidth]{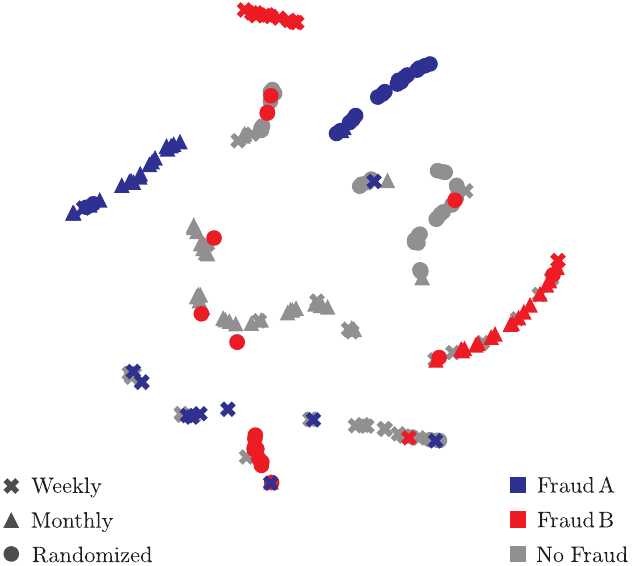}
    \vspace{0cm}
    \caption{t-SNE dimensionality reduction applied to latent representations $\vect{w}_{ij}$ of the different transaction set and fraud type combinations.}
    \label{fig:tsne}
\end{figure}

\emph{This section was omitted from the original publication in order to shorten the document.} To verify that the trained architectures really extract information from the transactions themselves, we inspect the convolutional filters learned by the $\Gamma$ learning function on the 1-hop synthetic financial transaction data set. We extract the learned parameters of the $\Gamma$ function from an instance of the best performing architecture (L4-GCN+), and initialize a stand-alone version of the mechanism. We generate a new set of transactions from the same distributions and feed them to the model in order to retrieve the latent representations. Next, we subject these embeddings to dimensionality reduction employing the t-SNE algorithm \citep{tsne}, the results of which can be seen in Figure \ref{fig:tsne}.

It is clear that the learning mechanism is able to offer latent representations based on which a distinction between the different transaction set and fraud types can be made. Similarly to other transaction set types, the majority of those of type 1 (weekly transactions) with fraud type \textbf{B} (represented by the red crosses \includegraphics[height=\fontcharht\font`\B]{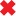}) appear in the same latent cluster. Interestingly, the embeddings offer a distinction between transaction type combinations beyond just the type of fraud, even though this was not required for the downstream task.

\begin{figure}[t]
    \centering
    \includegraphics[width=\columnwidth]{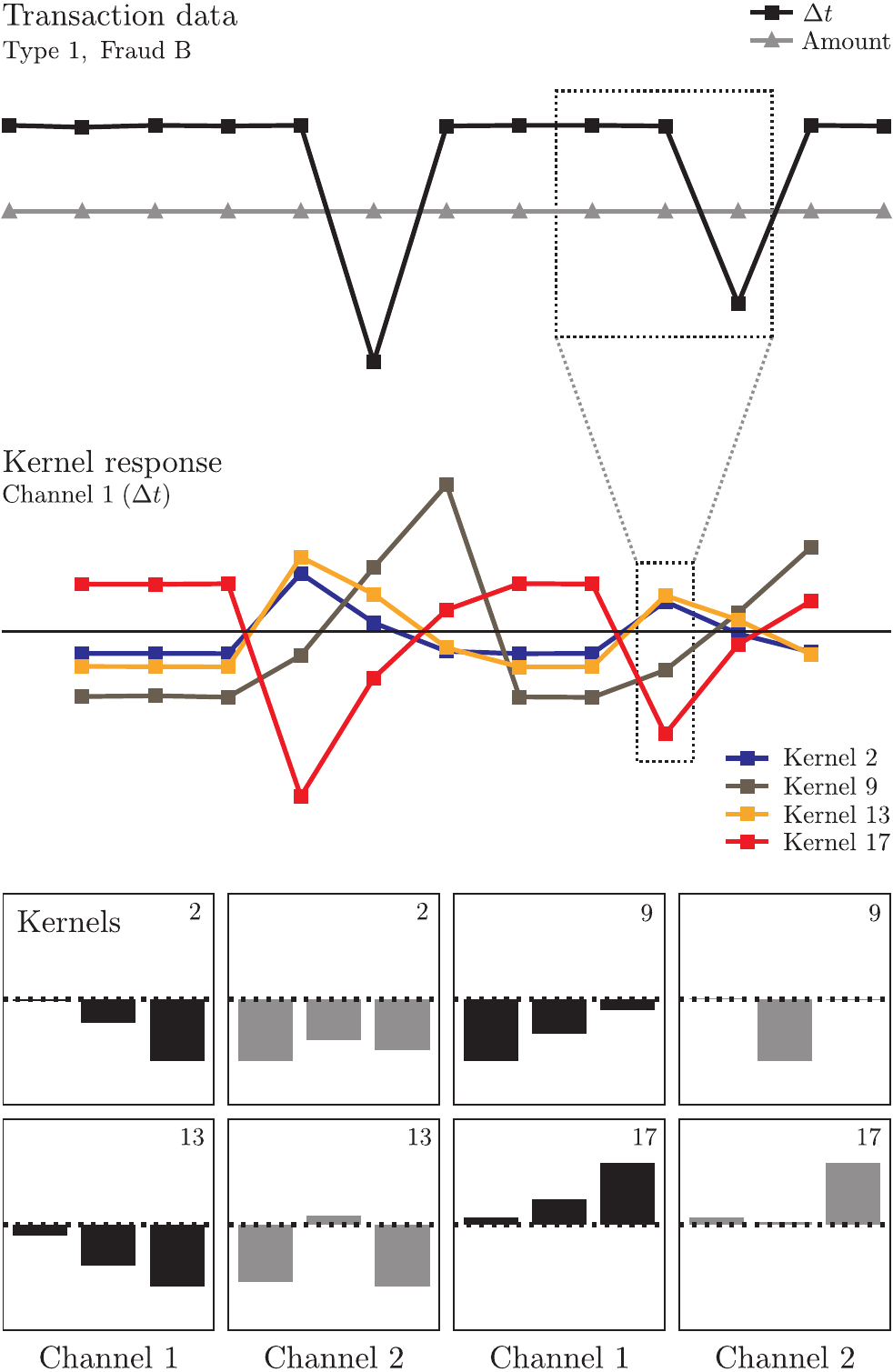}
    \caption{An example transaction set of type 1 and fraud type B (\textit{top}), convolutional kernels related to such patterns (\textit{bottom}) and their response to the data (\textit{middle}). Channel 2 responses are omitted since the input is constant. Kernel parameter values have been corrected for the kernel bias and are all scaled to within the same domain, since weights further down the architecture determine their influence.}
    \label{fig:kernels}
\end{figure}

We can delve deeper into this example by inspecting the $\Gamma$ learning mechanism at the level of the convolutional kernels, in order to identify the relevant filters. This can be seen in Figure \ref{fig:kernels}. Displayed are an example of a transaction set of type 1 and fraud type \textbf{B} (\textit{top}), convolutional kernels related to such patterns (\textit{bottom}) and their response to the data (\textit{middle}). Transaction sets of type 1 and fraud type \textbf{B} present themselves with sudden, one-off drops in the values in channel 1 ($\log \Delta t$), corresponding with occasional, chance-based double transactions (see Section \ref{subsubsection:transactionsets}). Some of the 20 kernels, 4 of which we show in Figure \ref{fig:kernels}, have learned typical Sobel-like filter configurations. These kernel structures are quintessentially related to edge detection, which is in alignment with the patterns in this type of transaction data, such as displayed in Figure \ref{fig:kernels}. We can conclude that with respect to our data sets, the $\Gamma$ learning mechanism is able to generate effective latent representations of the multi-edge populations by extracting patterns from the transaction data, indicating successful end-to-end learning.

\section{TRANSACTION NETWORK GENERATION}
\subsection{Fraud Detection}
We simulate the task of fraud detection in a network of financial transactions between corporate entities. The challenge is to separate the ``normal'' actors from the fraudulent. Class labels are only available at the node level, but relevant information also resides in the raw transaction data.

\subsection{Network Generation}
\label{subsubsection:networkgeneration}
We first generate a set of 50,000 vertices $v_i \in \mathcal{V}$, randomly allocated to one of two classes, \textbf{N} (normal) or \textbf{F} (fraud), with a class ratio of 9:1. Next, we generate 125,000 edges $e_{ij} \in \mathcal{E}$ between said vertices to create our graph. The order of magnitude of these numbers and the vertex-to-edge ratio are similar to those of data sets commonly used for testing GCN-like architectures, such as Citeseer, Cora and Pubmed (scientific citation networks) \citep{yang2016revisiting}.

Since most real-world networks have degree distributions that follow power laws \citep{ravasz2002hierarchical}, we employ the concept of \textit{preferential attachment} \citep{RevModPhys7447} in order to mimic this. In the original version of this model, one starts with a small, connected network, after which vertices are added one-by-one. The probability of a new node $v_i$ to be connected to one of the existing nodes $v_j$ scales directly with the number of edges the latter already has (its node degree $k_j$):
\begin{equation}
    p(e_{ij}) = \frac{k_j}{\sum_{\mathcal{V}} k}.
    \label{eq:preferentialattachment}
\end{equation}
Since all nodes already have been generated, we need to transform this expression into probabilities for each node $v_i$ to be either the source or the target node of a newly introduced edge:
\begin{equation}
    \label{eq:preferentialattachment2}
    p(v_i) \propto (k_i+1),
\end{equation}
the added +1 being required to allow new connections to zero-degree nodes. We shall denote the sampling probabilities for all $v_i$ by $P_\mathcal{V}$. This reformulation introduces control over the final number of edges, yet at the cost of concluding the generation process with a certain portion of zero-degree nodes and self-connections, both of which we opt to remove as the former are not part of the graph and the latter are unwanted remnants of the generation process. It is important to note that node degrees $k$ used in Equation \ref{eq:preferentialattachment2} are sums of in and out degrees. Based on these probabilities, we can now sample edges until a satisfactory number $|\mathcal{E}|$ is reached. For an overview of the generator, see Algorithm \ref{algo:graphgen}.

\begin{algorithm}
    \fontsize{9}{10}\selectfont
    $|\mathcal{V}|=50000$; $|\mathcal{E}|=125000$\;
    compute $P_\mathcal{V}$\;
    \While{$\sum_{i,j} \vect{A}_{ij} \leq |\mathcal{E}|$}{
        sample $(i,j)$ from $P_\mathcal{V}$\;
        $\vect{A}_{ij} \gets 1$\;
        recompute $P_\mathcal{V}$\;
    }
    \tcp{remove self-connections}
    \For{$e_{ij} \in \mathcal{E}$}{
        \uIf{$i = j$}{
            $\vect{A}_{ij} \gets 0$\;
        }
    }
    \tcp{remove zero-degree nodes}
    \For{$v_i \in \mathcal{V}$}{
        \uIf{$\sum_i \vect{A}_{ij} = 0$ {\bf and} $\sum_j \vect{A}_{ij} = 0$}{
            delete $v_i$ from $\mathcal{V}$\;
        }
    }
    \caption{Graph generator}
    \label{algo:graphgen}
\end{algorithm}

\subsection{Node Attributes}
All vertices receive a total of 13 features $\mathcal{F}$. 4 features represent typical properties of corporate entities: number of employees, turnover, profit and equity. 4 features are a one-hot encoding of industry sectors and the final 5 features are a one-hot encoding of regions of operation. Irrespective of their class, all node features are i.i.d. sampled from fictive distributions associated with these features. The two classes have the exact same distributions for each of the attributes. This should impede proper classification when looking at the node attributes exclusively. For an overview of the distributions used, see Section \ref{subsection:nodefeaturedists}.

\subsection{Transaction Sets}
\label{subsubsection:transactionsets}
\begin{figure*}
    \includegraphics[width=\textwidth]{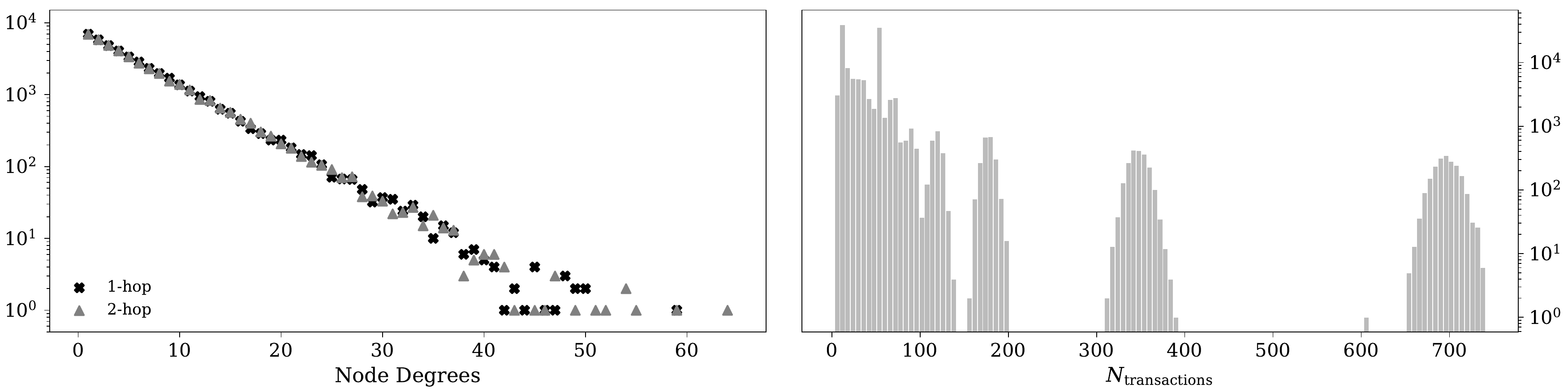}
    \caption{Node degree distribution (in + out) for both synthetic data sets (\textit{left}). Distribution of transaction set (edge populations) sizes for the 1-hop data set (\textit{right}). The distribution for the 2-hop data set is similar.}
    \label{fig:degree_transactions}
\end{figure*}

We can now add sets of transactions $\mathcal{S}_{ij}$ to the previously sampled edges. All transactions take place within the same time span of a year and have two attributes: time and amount. An edge can receive one of three types of \textit{transaction contracts}:
\begin{enumerate}
    \itemsep0em
    \item weekly payments of fixed amount,
    \item monthly payments of fixed amount,
    \item payments with random intervals and random amount.
\end{enumerate}
Types 1 and 2 receive additional small, randomized offsets on their time attributes in order to introduce a degree of noise. Depending on the classes of the source and target nodes, these transaction sets $\mathcal{S}_{ij}$ are modified. In case that $v_i \in \mathbf{F}$ and $v_j \in \mathbf{N}$ we introduce fraud type \textbf{A}, having the following effects (in order of the previously discussed transaction types)\footnote{All of these modifications take place with a per-transaction probability of $1/3$.}:
\begin{enumerate}
    \itemsep0em
    \item some weekly payments are missing,
    \item some monthly payments are missing,
    \item some payments have a decreased amount by a factor 10.
\end{enumerate}
In case $v_i \in \mathbf{N}$ and $v_j \in \mathbf{F}$ we introduce fraud type \textbf{B}, having similar but opposite effects:
\begin{enumerate}
    \itemsep0em
    \item some weekly payments occur twice,
    \item some monthly payments occur twice,
    \item some payments have an increased amount by a factor 5.
\end{enumerate}
For a summary of the transaction generator, see Algorithm \ref{algo:transgen}. The exact probability distributions can be found in Section \ref{subsection:transactiondists} and the generated distribution of transaction set sizes can be found in Figure \ref{fig:degree_transactions}.

\begin{algorithm}
    \fontsize{9}{10}\selectfont
    \For{$e_{ij} \in \mathcal{E}$}{
        sample $transType$ uniformly from $\{0,1,2\}$\;
        \uIf{$v_i \in \mathbf{F}$ {\bf and} $v_j \in \mathbf{N}$}{
            $fraudType \gets$ {\bf A}\;
        }
        \uElseIf{$v_i \in \mathbf{N}$ {\bf and} $v_j \in \mathbf{F}$}{
            $fraudType \gets$ {\bf B}\;
        }
        \uElse{
            $fraudType \gets$ None\;
        }
        $\mathcal{S}_{ij} \gets makeTransactions(transType,fraudType)$\;
    }
    \caption{Transaction set generator}
    \label{algo:transgen}
\end{algorithm}

\subsection{2-Hop Structure}
As outlined in the previous section, node classes have a direct effect on the characteristics of their incoming and outgoing multi-edge populations $\mathcal{S}_{ij}$. This means that most relevant information with respect to the classification task is located within 1 \textit{hop} from target vertices (their direct local neighborhood). In order to facilitate an interesting comparison, we therefore also generate a second synthetic data set, in which we introduce this correlation at a 2-hop distance.

After generation of the graph, we assign all direct neighbors of $v_i \in \mathbf{F}$ to an auxiliary ``mules'' class \textbf{M}. Instead of transaction sequences involving nodes of classes \textbf{F} and \textbf{N} (see Algorithm \ref{algo:transgen}), we now modify those involving nodes of classes \textbf{M} and \textbf{N} in a similar fashion. All $v_i \in \mathbf{M}$ are then reassigned to \textbf{N}. Note that this entails that \textbf{M} is a \textit{hidden class}, with $\mathbf{M \subseteq N}$: all \textbf{M} are labeled \textbf{N} in the final data set. The multi-edge populations $\mathcal{S}_{ij}$ modified based on their proximity to a vertex $v_i \in \mathbf{F}$, are now located 2 hops away from said vertex. This should pose a greater node classification challenge to our architectures, since information now is hidden at a greater distance within the data set.

\subsection{Data Summary}
An overview of the two data sets can be seen in Table \ref{table:datasets2}. Values for $|\mathcal{V}|$ differ from their initial values due to the removal of zero-degree nodes (see Section \ref{subsubsection:networkgeneration}). Values for $|\mathcal{E}|$ also differ slightly because of the removal of generated self-connections.

Before the data sets are passed to the architectures, all node features are normalized: within each attribute, data points are transformed linearly in order to fit inside a [0,1] range. In the transaction sets, individual times are transformed into relative time deltas within each sequence, disregarding all first transactions. Both the time deltas and transaction amounts are then transformed into their log values, with an appropriate normalization constant (again to produce data points approximately within a [0,1] range)\footnote{Our data sets and implementations can be found on GitHub: github.com/florishermsen/L-GCN}.

\begin{table}[h]
    \captionof{table}{The synthetic data sets and their characteristics: number of vertices and edges, sizes of classes \textbf{F}, \textbf{N} and \textbf{M}, and total number of transactions. Note that $\mathbf{M} \subseteq \mathbf{N}$, with no labels for $\mathbf{M}$ in the final data set.}
    \label{table:datasets2}
    \vspace{0.1cm}
    \fontsize{8}{9}\selectfont
    \begin{tabularx}{\columnwidth}{c@{ \ \ \ \ }c@{ \ \ }c@{ \ \ }c@{ \ \ }c@{ \ \ }c@{ \ \ }c}
          & $|\mathcal{V}|$ & $|\mathcal{E}|$ & $|\mathbf{N}|$ & $|\mathbf{F}|$ &  $|\mathbf{M}|$ & $\sum_\mathcal{E}|\mathcal{S}|$\\
        \hline
        \hline
        \textbf{1-hop} & 41792 & 124996 & 37574 & 4218 & - & 6643964\\
        \textbf{2-hop} & 41758 & 124995 & 37585 & 4173 & 11392 & 6618483\\
        \hline
    \end{tabularx}
\end{table}

\subsection{Transaction Data Distributions}
\label{subsection:transactiondists}

\subsubsection{Type 1}
The same amount for all transactions within the same sequence:
\begin{flalign*}
    P(\text{am}) \propto
        \begin{cases}
            e^\frac{-(\text{am}-30)^2}{2(5)^2}, & \text{if } \text{am} > 0 \\
            0,              & \text{otherwise}
        \end{cases}&&
\end{flalign*}
Different time deltas for individual transactions:
\begin{flalign*}
    \Delta t = 7 \text{\ days} + A \text{\ minutes}&&
\end{flalign*}
\begin{flalign*}
\vspace{0.1cm}
    P(A) \propto e^\frac{-(A)^2}{2(2)^2}&&
\end{flalign*}
\subsubsection{Type 2}
The same amount for all transactions within the same sequence:
\begin{flalign*}
    P(\text{am}) \propto
        \begin{cases}
            e^\frac{-(\text{am}-200)^2}{2(15)^2}, & \text{if } \text{am} > 0 \\
            0,              & \text{otherwise}
        \end{cases}&&
\end{flalign*}
Different time deltas for individual transactions:
\begin{flalign*}
    \Delta t = 30 \text{\ days} + B \text{\ minutes}&&
\end{flalign*}
\begin{flalign*}
    P(B) \propto e^\frac{-(B)^2}{2(2)^2}&&
\end{flalign*}

\subsubsection{Type 3}
Same maximum amount for all transactions within the same sequence:
\begin{flalign*}
    P(\text{max}) \propto
        \begin{cases}
            |e^\frac{-(\text{max}-220)^2}{2(100)^2}|, & \text{if } \text{T} \in \mathbb{Z} \\
            0,              & \text{otherwise}
        \end{cases}&&
\end{flalign*}
Different amounts for individual transactions:
\begin{flalign*}
    P(\text{am}) \propto
        \begin{cases}
            e^{-\text{am}/3000},& \text{if } 10 \leq \text{am} \leq \text{max} \\
            0,              & \text{otherwise}
        \end{cases}&&
\end{flalign*}
The same time delta base T for  all transactions within the same sequence:
\begin{flalign*}
    P(\text{T}) \propto
        \begin{cases}
            |e^\frac{-(\text{T}-10)^2}{2(10)^2}|, & \text{if } \text{T} \in \mathbb{Z} \\
            0,              & \text{otherwise}
        \end{cases}&&
\end{flalign*}
Different time deltas for individual transactions:
\begin{flalign*}
    \Delta t = T \text{\ days} + C \text{\ days} + D \text{\ hours} + E \text{\ minutes}&&
\end{flalign*}
\begin{flalign*}
    P(C) \propto e^\frac{-(T)^2}{2(T/2)^2}&&
\end{flalign*}
\begin{flalign*}
    P(D) \propto
        \begin{cases}
            1, & \text{if } D \in \{1,2, ... 24\} \\
            0,              & \text{otherwise}
        \end{cases}&&
\end{flalign*}

\begin{flalign*}
    P(E) \propto
        \begin{cases}
            1, & \text{if } E \in \{1,2, ... 60\} \\
            0,              & \text{otherwise}
        \end{cases}&&
\end{flalign*}

\subsection{Node Feature Distributions}
\label{subsection:nodefeaturedists}
\subsubsection{Number of Employees}

\begin{flalign*}
    P(x) \propto
        \begin{cases}
            e^{-0.005x_1},& \text{if } 10 \leq x_1 \leq 1500 \\
            0,              & \text{otherwise}
        \end{cases}&&
\end{flalign*}

\subsubsection{Turnover}

\begin{flalign*}
    P(x_2) \propto
        \begin{cases}
            e^{-0.00005x_2},& \text{if } 1\times10^{4} \leq x_2 \leq 1\times10^{7} \\
            0,              & \text{otherwise}
        \end{cases}&&
\end{flalign*}

\subsubsection{Profit}

\begin{flalign*}
    x_3 = x_2 - x_3'&&
\end{flalign*}
\begin{flalign*}
    P(x_3') \propto e^\frac{-(x_3')^2}{2(0.5x_3)^2}&&
\end{flalign*}

\subsubsection{Equity}

\begin{flalign*}
    P(x_4) \propto
        \begin{cases}
            e^{-0.00003x_4},& \text{if } 1\times10^{5} \leq x_4 \leq 1\times10^{7} \\
            0,              & \text{otherwise}
        \end{cases}&&
\end{flalign*}

\subsubsection{Sector}
\begin{flalign*}
    P(\text{S}) \propto
        \begin{cases}
            2+\sin{(\text{S})}^2,& \text{if } \text{S} \in \{0,1,2,3\} \\
            0,              & \text{otherwise}
        \end{cases}&&
\end{flalign*}

\subsubsection{Region}

\begin{flalign*}
    P(\text{R}) \propto
        \begin{cases}
            3+\sin{(\text{R+1})}^2,& \text{if } \text{R} \in \{0,1,2,3,4\} \\
            0,              & \text{otherwise}
        \end{cases}&&
\end{flalign*}

\end{document}